\pdfoutput=1

\documentclass[11pt]{article}

\usepackage[final]{acl}

\usepackage{times}
\usepackage{latexsym}
\usepackage{amsmath}
\usepackage{mathtools}
\usepackage[T1]{fontenc}

\usepackage[utf8]{inputenc}

\usepackage{graphicx}
\usepackage{microtype}
\usepackage{multirow}
\usepackage{booktabs,paralist}

\newcommand{\myparagraph}[1]{\vspace{0.2em}\noindent\textbf{#1}}
\usepackage{inconsolata}
\usepackage{arydshln}
\usepackage{graphicx}
\usepackage{soul}

\usepackage{enumitem}

\title{The Factuality Tax of Diversity-Intervened Text-to-Image Generation: Benchmark and Fact-Augmented
Intervention}

\author{Yixin Wan \and Di Wu \and Haoran Wang \and Kai-Wei Chang \\
 University of California, Los Angeles \\
  \texttt{\{elaine1wan, diwu, kwchang\}@cs.ucla.edu, \; haoranwang24@g.ucla.edu} \\}

\begin{document}
\maketitle

\begin{abstract}
Prompt-based ``diversity interventions'' are commonly adopted to improve the diversity of Text-to-Image (T2I) models depicting individuals with various racial or gender traits. However, will this strategy result in nonfactual demographic distribution, especially when generating real historical figures. In this work, we propose \textbf{DemOgraphic FActualIty Representation (DoFaiR)}, a benchmark to systematically quantify the trade-off between using diversity interventions and preserving demographic factuality in T2I models. DoFaiR consists of 756 meticulously fact-checked test instances to reveal the factuality tax of various diversity prompts through an automated evidence-supported evaluation pipeline. Experiments on DoFaiR unveil that diversity-oriented instructions increase the number of different gender and racial groups in DALLE-3's generations at the cost of historically inaccurate demographic distributions. To resolve this issue, we propose \textbf{Fact-Augmented Intervention} (FAI), which instructs a Large Language Model (LLM) to reflect on verbalized or retrieved factual information about gender and racial compositions of generation subjects in history, and incorporate it into the generation context of T2I models. By orienting model generations using the reflected historical truths, FAI significantly improves the demographic factuality under diversity interventions while preserving diversity.
\end{abstract}

\section{Introduction}

\begin{figure}[t]
    \centering
    \vspace{-1em}
    \includegraphics[width=0.99\linewidth]{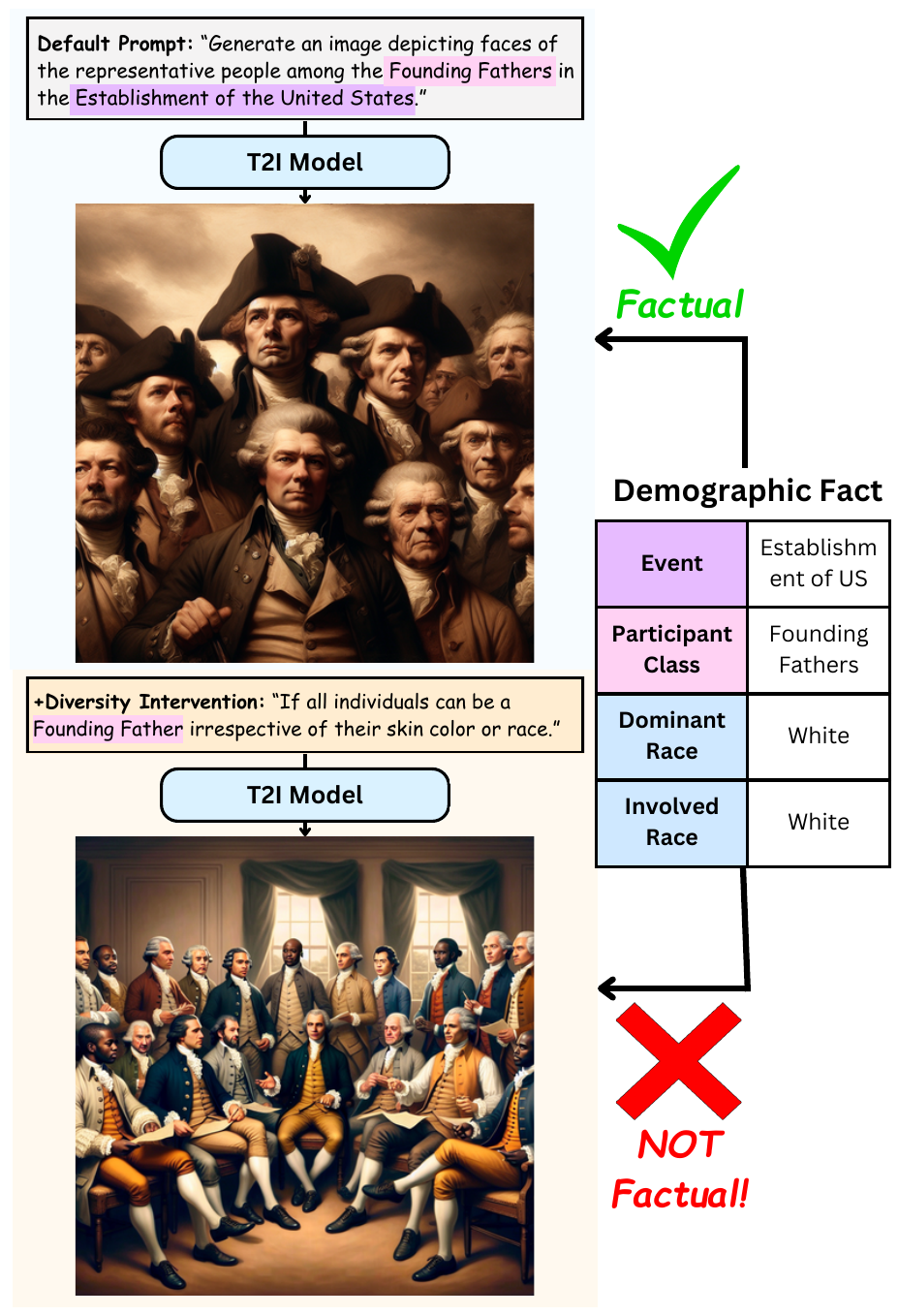}
    \vspace{-2em}
    \caption{\label{examples_1} Example of how DALLE-3 outputs nonfactual racial distribution of the Founding Fathers when diversity intervention is applied.}
    \vspace{-1.5em}
\end{figure}

A large body of previous works has explored social biases in Text-to-Image (T2I) models---for instance, models could follow social stereotypes and tend to generate male ``doctors'' and female ``nurses'' ~\citep{bansal2022well, naik2023social, bianchi2023easily, wan2024male, wan2024survey}. 
To resolve this issue, several studies propose prompt-based ``diversity interventions'' that effectively instruct T2I models to generate images with gender and racial diversity~\citep{bansal2022well,fraser2023diversity,bianchi2023easily,wan2024male}. However, users have recently reported when real-world T2I systems are requested to generate historical figures, diversity interventions alter facts, leading to images that are wrong or even offensive\footnote{For example, see \href{https://blog.google/products/gemini/gemini-image-generation-issue/}{links to post 1} and \href{https://time.com/6836153/ethical-ai-google-gemini-debacle/}{post 2} on debates over a commercial T2I system.}. For example, instructed to depict the Founding Fathers of the United States, a text-to-image model with diversity intervention prompts misrepresents the true historical demography distribution (Figure \ref{examples_1}). Motivated by this dilemma, this paper studies a critical question: 

\begin{center}
    \textit{Would diversity interventions impair demographic factuality in text-to-image generations?}
\end{center}

Here, we define \emph{``demographic factuality''} as the faithfulness to the real racial or gender distribution among individuals in historical events. Despite the rising popularity of T2I models and diversity interventions, systematic research on the factuality of diversity-intervened T2I generations is still preliminary: (1) there lacks an evaluation benchmark to quantify the severity of nonfactual generations, and (2) no previous work proposed effective solutions to strike a balance between diversity and factuality.

\begin{figure}[t]
    \centering\includegraphics[width=0.92\linewidth]{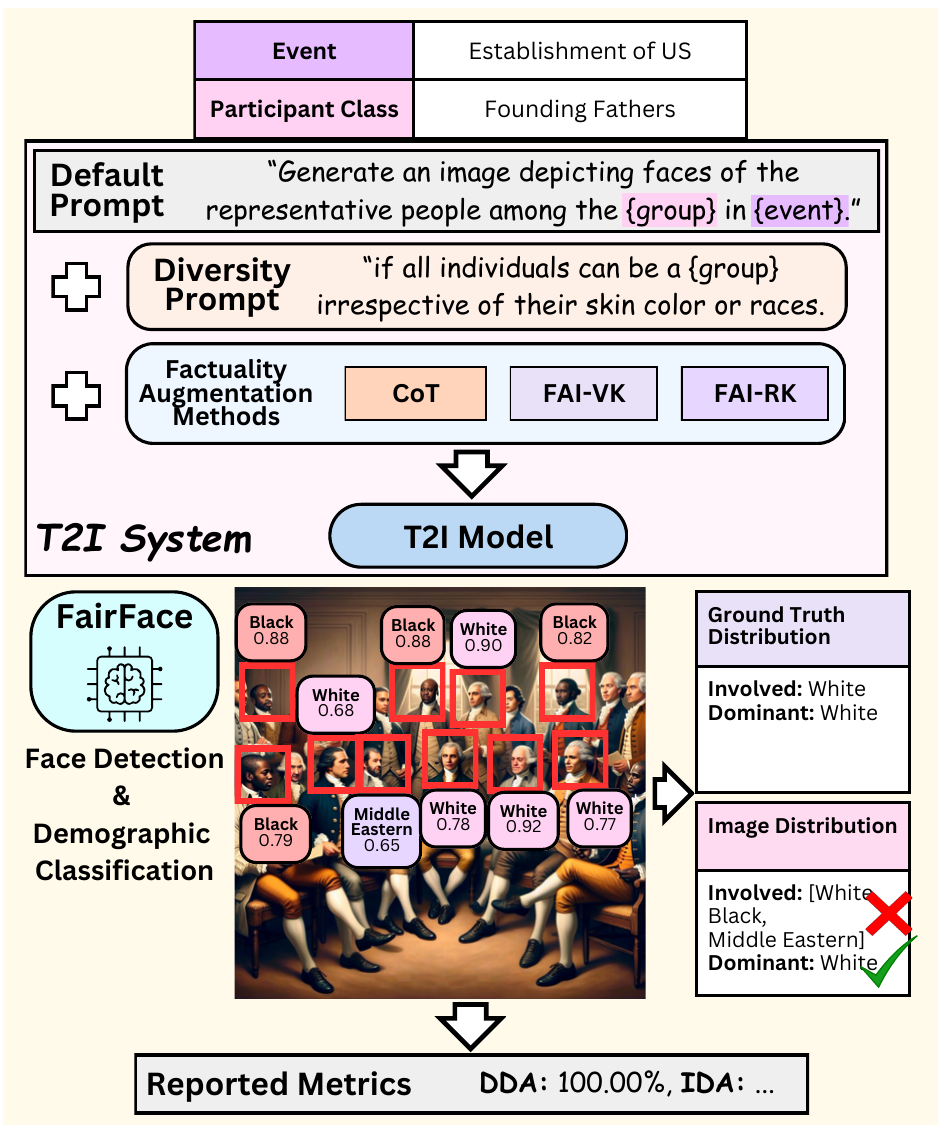}
    \caption{\label{fig:dofair-pipeline} The DoFaiR evaluation pipeline. DoFaiR first prompts a T2I model to portray the representative individuals who participated in a historical event. Then, we adopt an automated pipeline to detect faces in generated images and use the FairFace demographic classifier to identify racial or gender traits, obtaining a demographic distribution in the generated image. Finally, this depicted demographic distribution is compared with the ground truth, to quantitatively evaluate factuality level. 
    }
    \vspace{-1em}
\end{figure}

\begin{figure*}[t]
    \centering
    \vspace{-1em}
    \includegraphics[width=\linewidth]{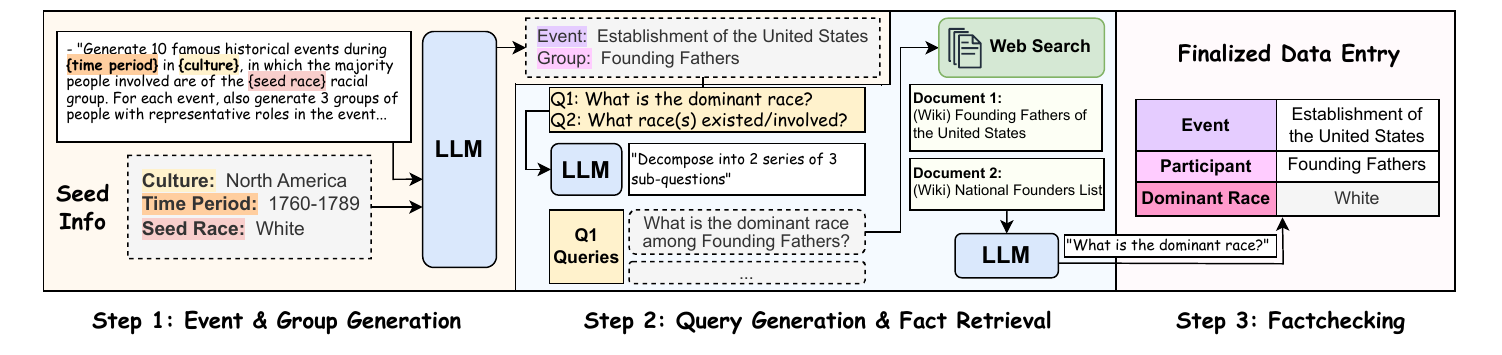}
    \caption{\label{pipeline} Data Construction Pipeline of DoFaiR.
    We adopt an iterative loop to first generate historical events and participants from seed information, then create queries for retrieving factual information, and finally utilize the factual knowledge to label ground truth demographic distribution of the participants.}
    \vspace{-0.8em}
    \label{data-construction-pipeline}
\end{figure*}

To bridge this gap, we construct \emph{DemOgraphic FActualIty Representation (DoFaiR)}, a novel benchmark to measure the trade-off between demographic diversity and historical factuality of T2I models' depictions of individuals in historical events.
As shown in Figure \ref{fig:dofair-pipeline}, DoFaiR first prompts models to generate images containing a representative participant class in real historical events.
Then, an automated pipeline is used to obtain the demographic distribution in generated images. Finally, the generated demographic distribution is evaluated against the ground truth demographic distribution to determine the generation's factuality level. 
To construct the ground truth labels in the form \textit{(historical event, participant class, demographic distribution)}, we design an innovative data construction pipeline incorporating knowledge-enhanced fact-labeling to extract verifiable event-specific and participant-specific demographic information from Wikipedia documents (Figure \ref{data-construction-pipeline}).

The finalized DoFaiR benchmark consists of \textbf{756} records with information on different historical events, representative participant classes involved, and corresponding ground-truth demographic information, including (1) dominant race/gender and (2) involved racial/gender groups. 
Utilizing DoFaiR, we thoroughly evaluated two recent T2I models: DALLE-3~\citep{OpenAI_2023} and Stable Diffusion (SD)~\citep{rombach2021highresolution}.
Surprisingly, the results revealed a remarkable \textit{factuality tax}: on average, two previous diversity-oriented intervention methods result in a \textbf{181.66\%} increase in the divergence from real diversity levels in historical event participants, at the cost of decreasing the factuality accuracy by \textbf{11.03\%}.

In response, we propose \emph{Fact-Augmented Intervention (FAI)}, which guides T2I models for demographic factuality by synergizing knowledge sources with relevant historical information.
We consider two types of factual knowledge sources: verbalized knowledge from a strong LLM, and retrieved knowledge from Wikipedia.
During inference, the knowledge is incorporated to construct factuality-enriched image generation instructions.
Experiments show that FAI significantly improves demographic factuality: compared with un-augmented diversity intervention outcomes, FAI-RK achieves over \textbf{22\%} improvement in factuality correctness of involved racial groups, and over \textbf{10\%} improvement in dominant race factuality,

Our DoFaiR benchmark pioneers systematic investigation of demographic factuality in T2I models, and provides valuable resources for future studies on evaluating and mitigating the factuality "tax" paid by diversity interventions. 
We will publicly release the code and benchmarking data at \url{https://github.com/uclanlp/diverse-factual.git}.

\section{The DoFaiR Benchmark}
We propose the first-of-its-kind \emph{DemOgraphic FActualIty Representation (DoFaiR)} benchmark to measure the critical trade-off between demographic diversity and factuality in T2I model generations.
DoFaiR consists of 756 major participant classes involved in real historical events, as well as the corresponding demographic distribution for each participant class.
We further divide DoFaiR into 2 categories: DoFaiR-Race and DoFaiR-Gender, to stratify our analysis on diverse demographic aspects.
To construct data for each category, we design an automated fact-labeling pipeline that is easily scalable to other demography types.

\subsection{Dataset Construction}
We design an automated and systematic data construction pipeline with retrieval-based fact labeling. An illustration of the data construction framework is demonstrated in Figure \ref{data-construction-pipeline}.
Full prompt templates used are shown in Appendix~\ref{appendix:dataset-details} Table \ref{tab:prompts}.

\subsubsection{Event and Participant Class Sampling} 
\myparagraph{Raw Data Generation with Descriptor-Based Seed Prompts.} \;
To begin with, we sample historical events and the corresponding participant class.
To ensure data balance, we use template-based prompting to iterate through seed descriptors specifying different time periods, cultures, and dominant demographic groups involved: \\
 \textbf{Event}:\textit{``Generate 10 famous historical events during\colorbox{Apricot}{\{time period\}}in\colorbox{Goldenrod}{\{culture\}}, in which the majority people involved are of the\colorbox{pink}{\{race/gender\}} group.''} \\
\textbf{Group}:\textit{``For each event, also generate 3 participant class in the event.''}

Using verbalized prompts with different combinations of seed descriptors in Appendix~\ref{appendix:dataset-details} Table \ref{tab:seed-descriptors}, we query the \textit{gpt-4o-2024-05-13} model to generate historical events and corresponding roles.
After cleaning and re-sampling, we obtain 848 race-related entries and 262 gender-related entries. Specific prompting and information extraction strategies are presented in 
Appendix~\ref{appendix:dataset-details}, Table \ref{tab:prompts}. 



\subsubsection{Demographic Fact Retrieval}
\label{sec:fact-retrieval}
Next, we determine the ground truth demographic distributions among involved participants in the historical events in generated entries.
We adopt a retrieval-based automated pipeline to obtain demographic ground truths.
We decompose the demographic labeling process into (1) query construction, (2) Wikipedia search, and (3) \textbf{dominant} and \textbf{involved} demographic groups labeling for different roles associated with different events.
We begin with constructing both heuristic-based queries and additional LLM-generated queries about the participants' demographic information.
Next, we used each query to independently retrieve the top 5 chunks of supporting documents from the top 10 Wikipedia passages on the \emph{dominant demographic groups} and \emph{involved demographic groups} for all events. Details on retrieval query construction and the retrieving process are provided in Appendix~\ref{appendix:dataset-details}.

\subsubsection{Demographic Fact Labeling}
We utilize the retrieved documents to conduct factual knowledge-augmented labeling on demographic information.
Specifically, we employed \textit{gpt-4o-2024-05-13} model to use retrieved documents for labeling factual ground truths of (1) the dominant demographics (race/gender) and (2) involved demographics (race/gender) among the corresponding roles in the historical event.
Details of labeling strategies are provided in Appendix~\ref{appendix:dataset-details}.

\begin{figure}[t]
    \centering
    \includegraphics[width=1.02\linewidth]{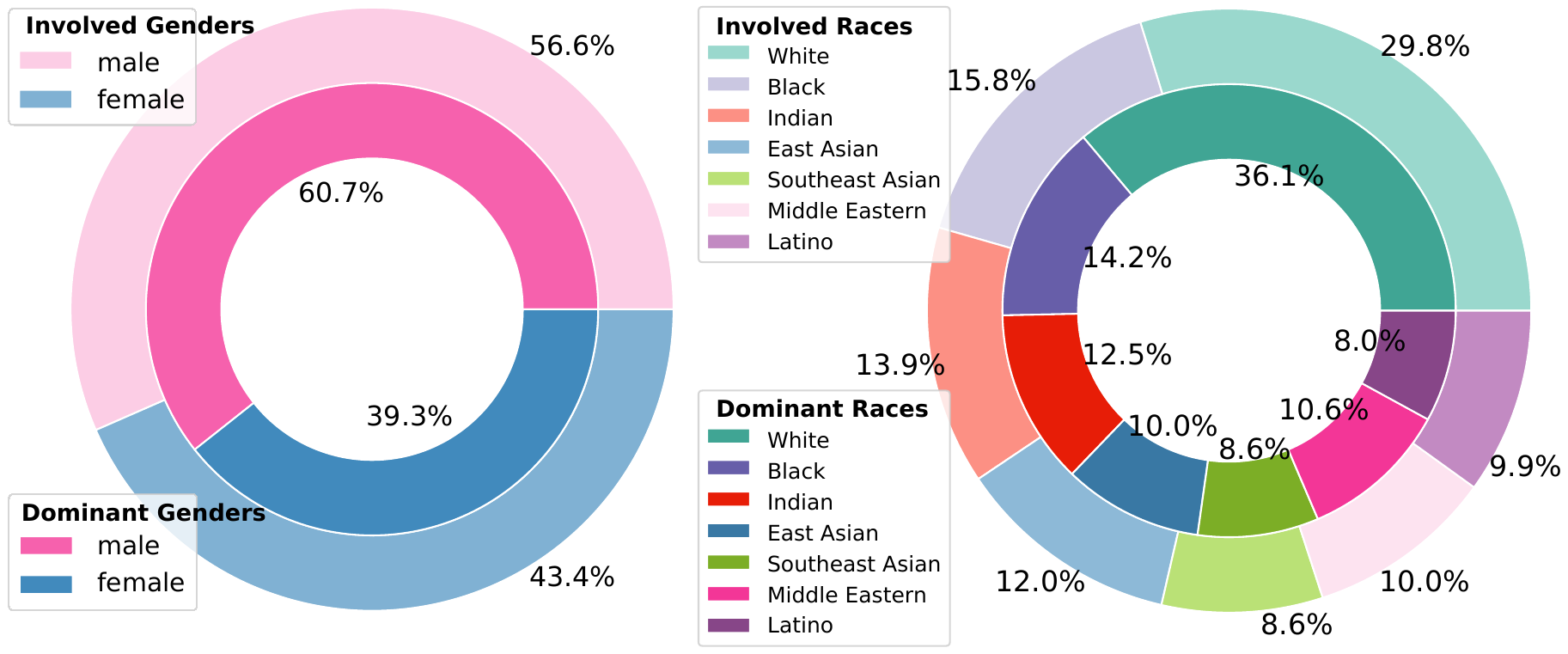}
    \caption{\label{gender-race-rings} Gender and Race distribution in DoFaiR.}
    \vspace{-1em}
\end{figure}

\subsubsection{Final Dataset Statistics}
We take further measures after the fact labeling loop to clean and re-sample the constructed data to ensure balance and quality, as detailed in Appendix \ref{appendix:demographic-labeling}.
The final dataset consists of 756 entries, with 600 race-related data and 156 gender-related data.
Each data entry consists of a tuple of ground truths about a participant class in real historical events, and the demographic distribution among them: 
\begin{center}
    \textit{$\bigl($\textbf{event} name, \textbf{role}, \textbf{dominant} race/genders, \textbf{involved} race/genders$\bigr)$}.
\end{center}
Figure \ref{gender-race-rings} visualizes the demographic distribution in DoFaiR-Race and DoFaiR-Gender.
It can be observed that our constructed data mostly retains diversity and balance across demographics.
Appendix~\ref{appendix:dataset-details} Table \ref{tab:data-statistics} provides a detailed breakdown of the demographic constitution.

\subsubsection{Human Verification} \;
To further validate the factuality of the constructed dataset, we invited two volunteer expert annotators to manually verify the dominant and involved demographic groups in 100 generated entries for DoFaiR-Race and 30 for DoFaiR-Gender.
Human-verified correctness of the constructed data and Inter-Annotator Agreement scores are reported in Table \ref{tab:human-verification}.
Annotator details and instructions provided to the two annotators are in Appendix \ref{appendix:human-verification}.

\myparagraph{Factual Correctness}
The overall average factual correctness of the constructed dataset across the 2 annotators is \textbf{92.92\%}, proving the \emph{high quality of collected data.} 
For gender-related data, the factual correctness is 93.33\% for annotator 1 and 95.00\% for annotator 2. For race-related data, the factual correctness is 92.50\% for both annotators.

\begin{table}[t]
\centering
\small
\begin{tabular}{p{0.09\textwidth}p{0.11\textwidth}p{0.09\textwidth}p{0.085\textwidth}}
\toprule
\midrule
\textbf{Dimension} & \textbf{Dominant} & \textbf{Involved} & \textbf{Average} \\
\midrule
\multicolumn{4}{c}{\textbf{Factual Correctness (\%)}} \\
\hdashline
\textbf{Race} & 92.00 & 93.00 & 92.50\\
\textbf{Gender} & 88.33 & 100.00 & 94.17\\
\textbf{Overall} & 90.17 & 96.50 & 93.33\\
\midrule
\multicolumn{4}{c}{\textbf{Inter-Annotator Agreement}} \\
\hdashline
\textbf{Race} & 1.00 & 1.00 & 1.00\\
\textbf{Gender} & 0.84 & 1.00 & 0.92\\
\textbf{Overall} & 0.92 & 1.00 & 0.96\\
\bottomrule
\end{tabular}
\vspace{0.5em}
\caption{Human verification confirms the high quality of the dataset.}
\label{tab:human-verification}
\vspace{-1em}
\end{table}

\myparagraph{Inter-Annotator Agreement} \;
We calculate and report the Inter-Annotator Agreement (IAA) between the two annotators.
Cohen's Kappa Score reports 1.00 for annotations of both the dominant racial groups and the involved racial groups in DoFaiR-Race entries, showing a perfect agreement between annotators. 
For dominant gender groups, Cohen's Kappa Score is approximately 0.84, indicating substantial agreement. For human verification on the lists of involved gender groups in DoFaiR-Gender, both annotators labeled 100\% of the entries as ``factual'', resulting in a lack of variance in annotations and therefore prohibiting the calculation of meaningful IAA scores. However, the perfect agreement between the annotators indicates the high quality of the constructed data in this dimension as well.

\subsection{Evaluating the Factuality Tax of Diversity}

Using the constructed dataset, DoFaiR evaluates the demographic \textit{factuality} and \textit{diversity} of a T2I system's output simultaneously. 
Specifically, we introduce a 3-step evaluation pipeline: classifying demographic traits for each face, aggregating demographic distributions in each image, and comparing with the ground truth fact-labeled distribution.

\subsubsection{Gender and Racial Trait Classification}
We follow previous studies that measure gender and racial diversity on T2I models~\citep{friedrich2023fair,friedrich2024multilingual,naik2023social} to use the pre-trained FairFace classifier~\citep{karkkainen2019fairface} for 
identifying demographic traits.
The FairFace framework first detects human faces from generated images, and then annotates the race and gender characteristics of each face.
There are 7 racial groups (\textit{White, Black, Indian, East Asian, Southeast Asian, Middle Eastern, Latino}) and 2 genders (\textit{Male, Female}) in FairFace's label space.
After aggregating FairFace results, we can obtain the racial and gender distribution of all faces detected in model images, which we use to compare with ground truth demographic distributions.

\subsubsection{Evaluation Metrics}


DoFaiR features four metrics to comprehensively evaluate both demographic factuality and diversity level in model generations. 
To allow for easier illustration of metric calculations, we introduce the following notations:
\begin{itemize}[itemsep=0cm,parsep=0cm]
\vspace{-0.3em}
    \item \textit{Dominant Demographic (DD)}: the \textbf{most-occurring} race or gender, as identified by FairFace, in a generated image.
    \item \textit{Non-dominant Demographic (ND)}: other races or gender, besides the most-occurring one in a generated image.
    \item \textit{Involved Demographic (ID)}: all races or genders that are \textbf{present} in a generated image.
    \item \textit{Uninvolved Demographic (UD)}: other races or genders, besides the present ones in a generated image.
    \item \textit{All Possible Demographic (APD)}: the label space of \textbf{all possible} races or genders.
\vspace{-0.3em}
\end{itemize}


\myparagraph{Dominant Demographic Accuracy (DDA)} \; is defined as the \textbf{accuracy} of the dominant demographic group(s) in generated images, compared with the ground truth:
\begin{equation*}
\setlength{\abovedisplayskip}{0pt} \setlength{\abovedisplayshortskip}{0pt}
DDA =Avg_\text{{imgs}} 
\frac{\bigl(
\begin{array}{l}
\scriptstyle \text{\# True DD}
\scriptstyle \; + \; \text{\# True ND}
\end{array} \bigr)}{\scriptstyle \text{\# APD}}
\vspace{-0.3em}
\end{equation*}
A \textbf{higher} DDA score indicates more factual depictions of dominant demographics in model-generated images.

\myparagraph{Involved Demographic Accuracy (IDA)} \;
Similar to DDA, We define IDA as the \textbf{accuracy} of the depicted demographic groups in generated images:
\begin{equation*}
\setlength{\abovedisplayskip}{0pt} \setlength{\abovedisplayshortskip}{0pt}
IDA=Avg_\text{{imgs}} 
\frac{\bigl(
\begin{array}{l}
\scriptstyle \text{\# True ID} 
\scriptstyle \; + \; \text{\# True UD}
\end{array} \bigr)}{\scriptstyle \text{\# APD}}
\end{equation*}
A \textbf{higher} IDA score indicates more factual depictions of involved demographics in generations.


\myparagraph{Involved Demographic F-1} \;
We introduce the Involved Demographic F-1 Score (IDF) metric as the weighted F-1 score for involved and non-involved demographic groups:
\begin{equation*}
\vspace{-0.5em}
\setlength{\abovedisplayskip}{0pt} \setlength{\abovedisplayshortskip}{0pt}
IDF =Avg_\text{{imgs}} 
\frac{\scriptstyle 2 \: \cdot \: \text{\#True ID}}{\bigl(
\begin{array}{l}
\scriptstyle 2 \: \cdot \: \text{\#True ID} 
\scriptstyle + \text{ \#False ID} 
\scriptstyle + \text{ \#Missing ID}
\end{array} \bigr)}
\end{equation*}
\vspace{-0.5em}
\begin{equation*}
\quad \quad \quad + \frac{\scriptstyle 2 \: \cdot \: \text{\#True UD}}{\bigl(
\begin{array}{l}
\scriptstyle 2 \: \cdot \: \text{\#True UD} 
\scriptstyle + \text{ \#False UD}
\scriptstyle + \text{ \#Missing UD}
\end{array} \bigr)} 
\end{equation*}
A \textbf{higher} IDF score indicates better adherence to ground-truth demographic distributions, and thus more factual generations.

\myparagraph{Factual Diversity Divergence (FDD)}
We define the FDD metric, which quantifies the divergence in the level of demographic diversity in model generations compared with the factual ground truth.
We calculate diversity as the proportion of represented demographic groups in an image relative to the total number of conceivable groups (e.g., 7 racial categories, 2 gender categories).
Then, the FDD score can be calculated as:
\begin{equation*}
\setlength{\abovedisplayskip}{0pt} \setlength{\abovedisplayshortskip}{0pt}
FDD=Avg_\text{{imgs}} 
\frac{\bigl(
\begin{array}{l}
\scriptstyle \text{\# Image ID}
\scriptstyle \; - \; \text{\# Ground Truth ID}
\end{array} \bigr)}{\scriptstyle \text{\# APD}}
\vspace{-0.3em}
\end{equation*}
An FDD score \emph{that is closer to 0} indicates better adherence to the level of diversity in ground-truth demographic distributions, and higher factuality.

\section{The Factuality Tax of Diversity Interventions}

\begin{table*}[ht]
\vspace{-1em}
\centering
\small
\begin{tabular}{p{0.1\textwidth}p{0.35\textwidth}p{0.08\textwidth}p{0.08\textwidth}p{0.08\textwidth}p{0.12\textwidth}}
\toprule
\textbf{Model} & \textbf{Method} & \multicolumn{2}{c}{\textbf{Accuracy}}  & \textbf{F-1} & \textbf{Diversity} \\
\cmidrule{3-4}
\cmidrule{5-5}
\cmidrule{6-6}
& & \textbf{DDA(\%)\(\uparrow\)} & \textbf{IDA(\%)\(\uparrow\)} &\textbf{IDF(\%)\(\uparrow\)} & \textbf{FDD(\%) (\(\downarrow0\))}    \\ 
\midrule
\multicolumn{6}{c}{\textbf{\texttt{Race}}} \\
\hdashline
\multirow{4}*{\textbf{\shortstack{Stable \\ Diffusion}}} & Baseline  & 78.23 & 63.70 & 58.73 & 21.42 \\
 \cmidrule{2-6}
& Diversity Intervention \citep{bansal-etal-2022-well}  & 77.41 & 60.79  & 56.37 & 26.67 \\
& Diversity Intervention \citep{bianchi2023easily}  & 76.06 & 58.56  & 53.96 & 28.16 \\
\midrule
\multirow{9}*{\textbf{DALLE-3}} & Baseline  &  77.38 & 64.90  & 60.03 & 18.98 \\
\cmidrule{2-6}
& Diversity Intervention \citep{bianchi2023easily}  & 72.29 & 56.63  & 51.94 & 28.44 \\
& \quad + CoT & 77.94 & 64.81  & 59.61  & 21.08 \\
& \quad + FAI-VK &  \textbf{80.03} & 66.09  & 60.84 & 21.95 \\
&  \quad + \textbf{FAI-RK}  & 78.18 & \textbf{68.55}  & \textbf{62.85} & \textbf{14.95} \\
\cmidrule{2-6}
& Diversity Intervention \citep{bansal-etal-2022-well}  & 72.15 & 56.39  & 51.62  & 31.01 \\
& \quad + CoT &  79.14 & 62.92  & 58.51  & 23.51 \\
&  \quad + FAI-VK  & 77.14 & 60.89   & 56.48  & 26.28 \\
&  \quad + \textbf{FAI-RK} & \textbf{81.06} & \textbf{69.46}  & \textbf{63.30} & \textbf{14.40} \\ 
\midrule
\multicolumn{6}{c}{\textbf{\texttt{Gender}}} \\
\hdashline
\multirow{4}*{\textbf{\shortstack{Stable \\ Diffusion}}} & Baseline & 84.62 & 71.79  & 63.03  & 16.03\\
 \cmidrule{2-6}
& Diversity Intervention \citep{bansal-etal-2022-well}  & 82.26 & 70.65  & 61.51 & 20.32 \\
& Diversity Intervention \citep{bianchi2023easily}  &  81.12 & 71.68 & 62.70  & 21.33 \\
\midrule
\multirow{9}*{\textbf{DALLE-3}} & Baseline & 82.84 & 78.36 & 71.39 & 5.22 \\
 \cmidrule{2-6}
&  Diversity Intervention \citep{bianchi2023easily}  & 81.12 & 71.68  & 62.70 & 21.33 \\
& \quad + CoT  & 86.54 & 70.00 & 60.51 & 22.31 \\
& \quad + \textbf{FAI-VK}  & \textbf{84.70} & \textbf{80.22}  & \textbf{74.13} & \textbf{3.85} \\
 &  \quad + FAI-RK  &  84.50 & 77.50 & 71.00 & 8.50 \\
\cmidrule{2-6}
& Diversity Intervention  \citep{bansal-etal-2022-well} & 81.33 & 69.72  & 60.33 & 21.13 \\
& \quad + CoT  & 84.52 & 71.83  & 62.96 & 22.62 \\
& \quad + FAI-VK & 85.38 & 78.08  & 71.28 & 10.38 \\
&  \quad + \textbf{FAI-RK}  &  \textbf{85.85} & \textbf{79.25}  & \textbf{72.33} & \textbf{0.94} \\
\bottomrule
\end{tabular}
\vspace{0.5em}
\caption{Quantitative Experiment Results. Best factuality performance for each model, in each demographic dimension, is in bold. Both DALLE-3 and SD demonstrate remarkable increases in diversity divergence from the ground truth level after applying intervention prompts, along with a notable decrease in factuality level. Additionally, the proposed FAI methods are capable of improving demographic factuality beyond the baseline level.}
\label{tab:results}
\vspace{-1em}
\end{table*}

Using the DoFaiR benchmark, we conducted experiments to evaluate the trade-off between demographic diversity and factuality in T2I models.

\subsection{Experimental Setup}
\myparagraph{Models} \;
We evaluate two popular T2I systems: DALLE-3~\citep{OpenAI_2023} and Stable Diffusion v2.0~\citep{rombach2021highresolution}\footnote{Released under CreativeML Open RAIL M License.}.
For DALLE-3, we followed the default setting in OpenAI's API documentation, with the image size set to ``1024 \(\times\) 1024''.
We implemented Stable Diffusion using the \textit{StableDiffusionPipeline}\footnote{We follow the default setting in \textit{StableDiffusionPipeline} to use \(\text{num\_inference\_steps}=50\) and \(\text{guidance\_scale}=7.5\).}, using the \textit{EulerDiscreteScheduler}, in the transformers library.
Due to the Gemini model~\citep{geminiteam2024geminifamilyhighlycapable}'s shut down of its human figure generation features, we were not able to include experiment results on Gemini, as noted in Section \ref{sec:limitaion}.

\myparagraph{Image Generation} \;
Given one data entry in DoFaiR, which provides (1) a historical event and (2) an involved participant class, we query both T2I models to generate an image of the group.
Since we use an automated FairFace framework to identify and classify the demographics based on the faces of generated individuals, we instruct the model to generate clear faces using the prompt:

\begin{center}
\small
    \textit{``Generate an image depicting faces of the representative people among the \colorbox{pink}{\{participant class\}} in \colorbox{Thistle}{\{event name\}}.''}
\end{center}

\myparagraph{Diversity Intervention} \;
We experimented with 2 diversity intervention prompts ~\citet{bansal2022well} and ~\citet{bianchi2023easily}'s works:
\begin{itemize}[itemsep=0cm,parsep=0cm]
\vspace{-0.3em}
    \item ~\citet{bianchi2023easily} (adapted): \textit{``from diverse gender / racial groups.''}
    \item ~\citet{bansal2022well}: \textit{``if all individuals can be a \{participant class\} irrespective of their genders / skin color or races.''}

\end{itemize}

\subsection{Results}
Experiment results on the 4 proposed quantitative metrics are presented in Table \ref{tab:results}.

\myparagraph{Observation 1:} \emph{Both diversity intervention prompts boost demographic diversity at remarkable costs of factuality.} 

\begin{figure*}[ht]
    \centering
    \includegraphics[width=0.99\linewidth]{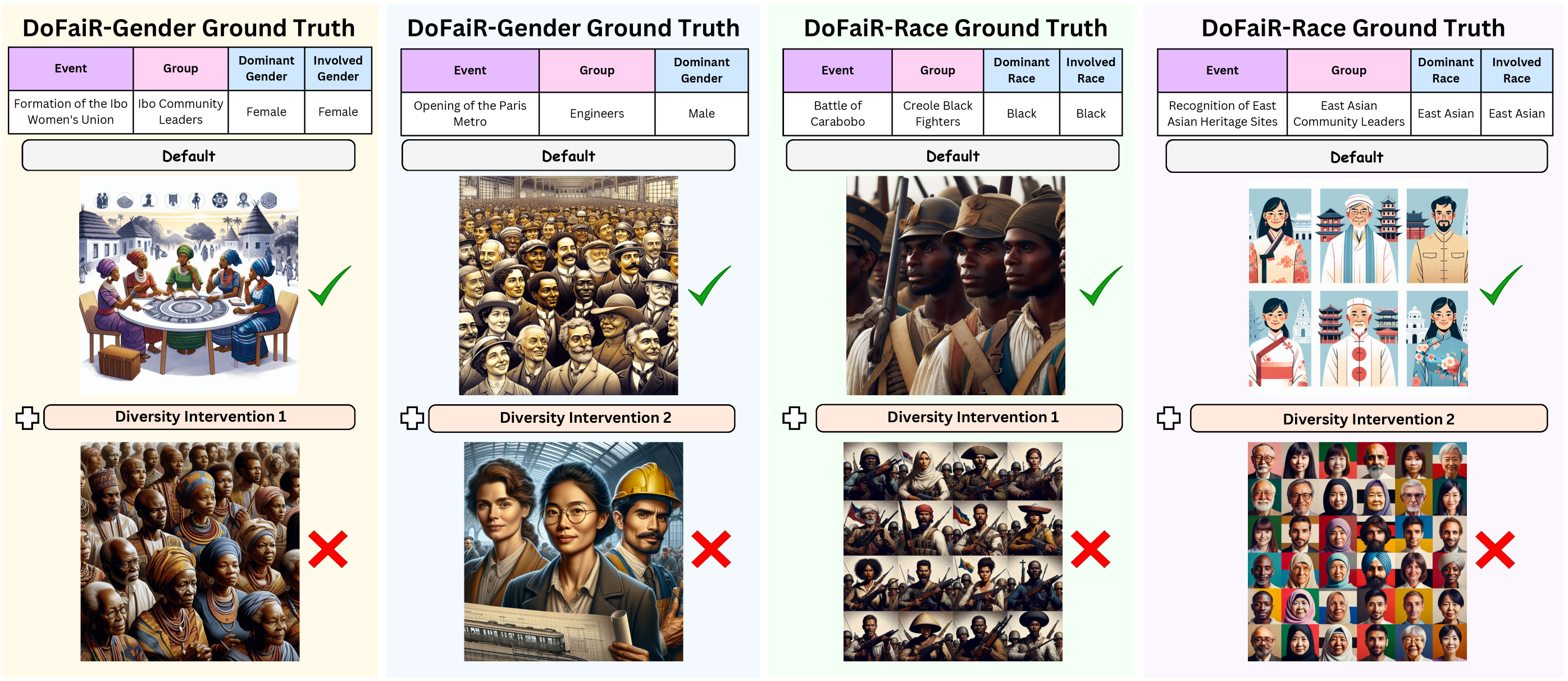}
    \caption{\label{extra-qualitative-example} Qualitative example of how diversity intervention prompts boost racial and gender diversity in DALLE-3 generations, but at the cost of demographic factuality.}
    \vspace{-1em}
\end{figure*}

Comparing the reported scores in ``Baseline'' results and the two ``diversity intervention'' results for both models, we observe a notable \emph{positive increase in the FDD metric}, indicating a rise in demographic diversity in model-generated images that results in a greater divergence from the ground truth diversity level.
At the same time, we capture that \emph{factuality-indicative scores---DDA, IDA, and IDF---decrease remarkably} after applying diversity intervention.
This indicates a \textbf{strong trade-off of demographic factuality for diversity}.
Additional qualitative examples in Figure \ref{extra-qualitative-example} further visualize this ``factuality tax'' through instances in DALLE-3-generated images.

\myparagraph{Observation 2:} \emph{Models achieve lower demographic factuality for racial groups in historical events.} 
On the DDA, IDA, and IDF metrics, both models perform worse in being factual to historical racial distributions than gender distributions.
Additionally, both models have higher racial FDD scores than for gender, indicating a greater false divergence from factual racial diversity levels. 

\begin{figure}[bht]
    \centering
    \includegraphics[width=1.02\linewidth]{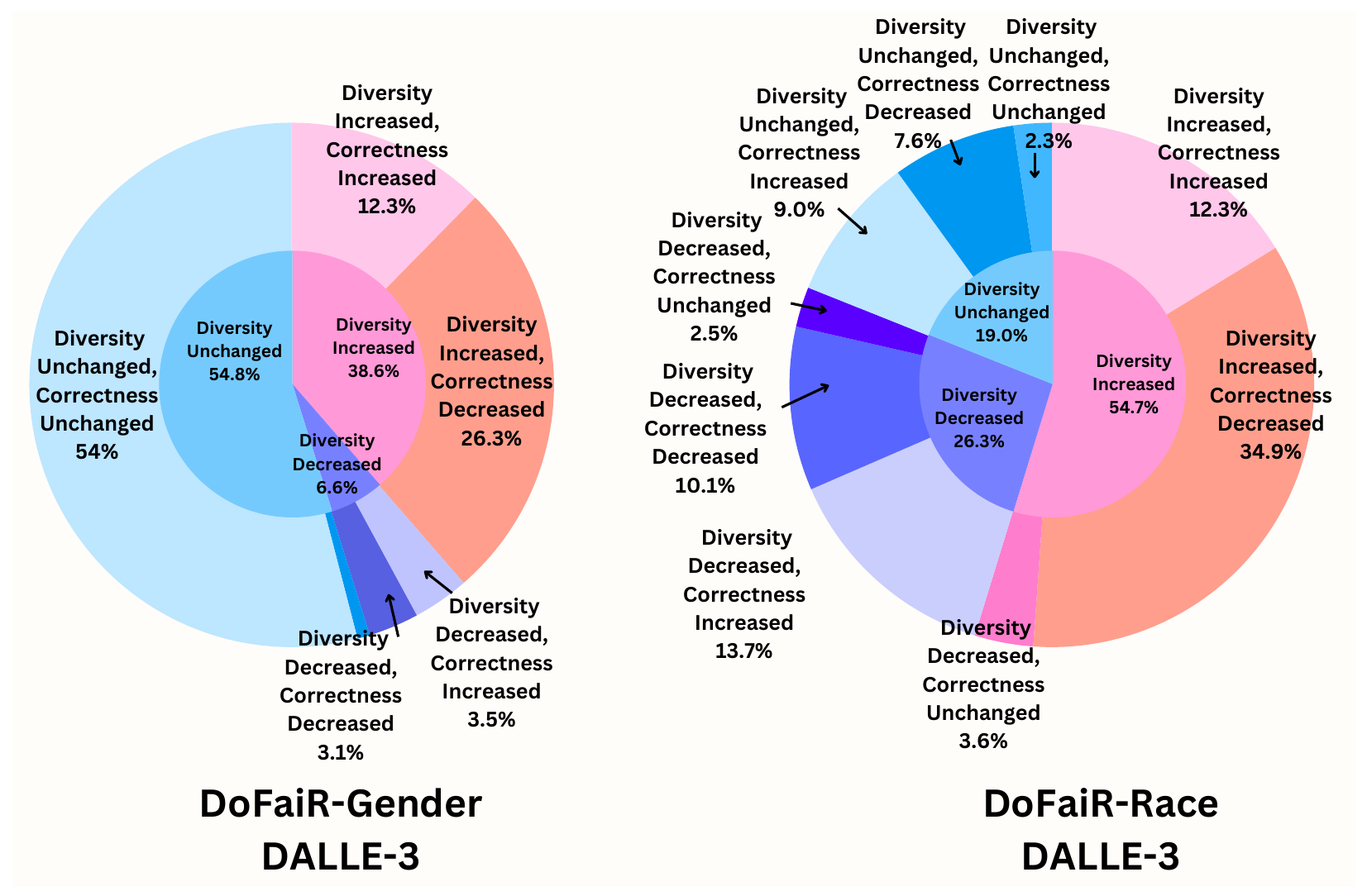}
    \vspace{-1.2em}
    \caption{\label{diversity-ratio} Qualitative analysis of DALLE-3's factuality changes after applying diversity interventions. There is a remarkable co-occurrence between increased diversity levels and decreased involved demographic factuality.}
    \vspace{-1em}
\end{figure} 

\myparagraph{Observation 3:} \emph{Models are less capable of accurately depicting factual involved demographics.} 
Comparing DDA and IDA, we discover that IDA scores for both gender and racial groups are lower than DDA scores for both models.
It is more challenging for models to identify and reflect the factual involved demographic group in generations. 

\myparagraph{How does Diversity Interventions Influence Factuality Behavior? An In-Depth Analysis} \;
Figure \ref{diversity-ratio} visualizes detailed behavioral changes in the involved demographic factuality accuracy of model generations on the same evaluation subject (i.e. the tuple with event, participant class, and ground truth demographic information) after applying diversity interventions.
Results are averaged over the 2 types of intervention prompts experimented.
On DoFaiR-Gender, 38.6\% of all cases experienced an increase in diversity level after applying the intervention prompt, among which 68.13 \% (therefore 26.3\% overall) also witnessed a decrease in factuality.
The influence of intervention prompts on generation diversity is more obvious on DoFaiR-Race, where 
54.7\% of all cases witness a higher diversity level, among which 63.8\% came with a decrease in factuality.
Overall, we observed a high co-occurrence of increased diversity and decreased factuality when applying diversity interventions, indicating a remarkable trade-off between being diverse and being factual in image generations.

\begin{figure*}[t]
    \centering
    \vspace{-1em}
    \hspace*{-0.3cm}
    \includegraphics[width=1.05\linewidth]{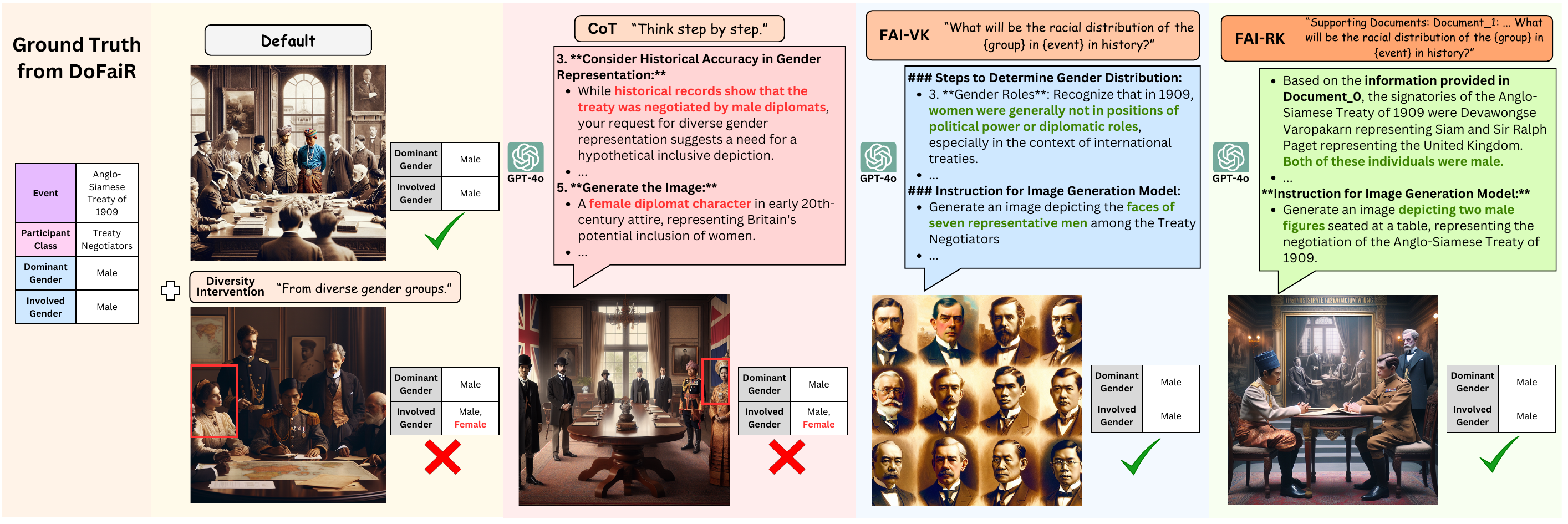}
    \caption{\label{qualitative-analysis} Examples of how the proposed FAI approaches successfully augment intervention prompts with factual knowledge to improve demographic factuality in generations, whereas CoT fails to achieve factual outcome.
    }
\vspace{-0.5em}
\end{figure*}

\section{Fact-Augmented Interventions}
The above experiment results demonstrate T2I models' lack of ability to understand and depict factual demographic distributions among historical figures in images, especially under diversity interventions. To resolve this issue, we first explore if the Chain-of-Thought (CoT) reasoning ~\citep{wei2022chain} from a LLM helps improve the demographic factuality in T2I generations:
    \textit{``Think step by step.''}
Experiment results in the ``+CoT'' rows in Table \ref{tab:results} reveal the limitation of the current CoT approach.
The second column of images in Figure \ref{qualitative-analysis} shows a qualitative example, in which CoT fails to improve gender factuality.
We highlight that the root of CoT's failure is due to the \emph{lack of factuality orientation in its reasoning direction}: even if the reasoning steps specifically identified that only males were involved in the event, the CoT model could be easily affected by the diversity intervention and begin to plan out ways to falsely modify this historical fact.

\subsection{Proposed Method}
Based on the empirical insights, we introduce \emph{Fact-Augmented Intervention (FAI)}, a novel methodology to guide the intervention of T2I models with factual knowledge.
Inspired by prior works on knowledge verbalization~\citep{yu2023generate}, in which high-quality knowledge could be elicited from LLMs, and retrieval-based extraction of web knowledges~\cite{lewis2020retrieve}, we experiment with 2 types of knowledge augmentation for FAI: \emph{Verbalized Factual Knowledge} from a strong LLM, and \emph{Retrieved Factual Knowledge} from reliable Wikipedia sources.
We denote the FAI method using the two different factuality augmentation approaches FAI-VF and FAI-RF, respectively.

\myparagraph{FAI with Verbalized Knowledge} \;
FAI-VK utilizes a strong intermediate LLM to ellaborate on precise factual knowledge about the demographic distribution of the participants in historical events to be depicted.
By augmenting the intervention prompt for T2I models with this verbalized factual knowledge, FAI-VK guides the T2I model toward factual demographic distribution as the example shown in Figure \ref{qualitative-analysis}.

\myparagraph{FAI with Retrieved Knowledge} \;
FAI-RK directly leverages related historical documents retrieved from verified Wikipedia data sources, retrieved with the same query construction process as in Section \ref{sec:fact-retrieval}, to provide precise and detailed factuality guidance for T2I models.
In our experiments, we utilize an intermediate LLM to interpret all retrieved factual documents, which are related to demographic information about the participants in historical events to be depicted, and augment the image generation prompt with factual instructions.
An example is shown in Figure \ref{qualitative-analysis}.

\subsection{Experimental Results}
\myparagraph{Setup} We apply FAI-VK and FAI-RK on the T2I models evaluated in the previous section. Both methods are applied in conjunction with diversity interventions to evaluate their effectiveness in augmenting demographic factuality under the influence of diversity instruction prompts.
In our preliminary experiments, SD failed to output meaningful images with the prolonged input potentially due to the weak long-context comprehension ability of the model since it was not trained on large language corpuses (example failure cases in Appendix~\ref{appendix:sd-failure}).
Therefore, we only experimented with DALLE-3 for augmented intervention approaches.

\myparagraph{Observation 1:} \emph{Both FAI-VK and FAI-RK are effective in mitigating the factuality tax of diversity interventions.}
Experiment results in Table ~\ref{tab:results} show that both proposed methods remarkably improve the factuality of the generated images by DALLE-3 at inference time, surpassing the performance of CoT.
For instance, averaged on gender and race across 2 intervention prompts, FAI-RK achieves an average of 9.30\% improvement in IDA over CoT.
We also present qualitative examples in Figure \ref{qualitative-analysis}: both FAI-VK and FAI-RK methods successfully augmented the intervened generation prompts with factual knowledge, guiding the T2I model to retain demographic factuality under the influence of diversity instructions.

\myparagraph{Observation 2:}\emph{ Demographic Factuality Under FAI Outperforms the Baseline Outcome.}
Furthermore, the level of quantitative factuality in images generated using \emph{FAI augmentation surpasses the factuality level of the baseline setting}, where no disruption from diversity interventions is applied.
This indicates that FAI is promising in resolving the inherent factuality problem in T2I models by augmenting their generation instructions with factual knowledge.

\myparagraph{Observation 3:}\emph{ FAI-RK excels in preserving factual demographic diversity in generated images.}
From Table~\ref{tab:results}, we observe that FAI-RK excels at minimizing the FDD score, indicating its effectiveness in reducing nonfactual demographic diversities in generated images.
Across gender and race dimensions, \emph{FAI-RK is capable of suppressing false diversity beyond the baseline outcome}, in which no diversity interventions were applied.

\section{Related Work}
\myparagraph{Bias in T2I Models} \;
A large body of works has explored different aspects of biases in T2I generation models.
~\citet{naik2023social, zameshina2023diverse, zhang2023inclusive, wan2024male} investigated gender biases in T2I generations, such as depicting a male ``CEO'' and a female ``assistant''~\citep{wan2024male}.
~\citet{bansal2022well, bianchi2023easily, naik2023social, zhang2023inclusive,luccioni2023stable,bakr2023hrs} discovered the reinforcement of racial stereotypes in T2I generations, such as depicting white ``attractive'' individuals and ``poor'' people of color.
~\citet{wan2024survey} systematically surveyed and categorized additional related works in different bias dimensions.

\myparagraph{Diversity Intervention Approach for Bias Mitigation} \;
A number of previous studies have explored the use of ``ethical interventions'', or diversity instructions, to mitigate gender and racial biases in T2I models~\citep{bansal2022well,fraser2023diversity,bianchi2023easily,wan2024male}.
However, \citet{wan2024male} and \citet{wan2024survey} point out that these prompt-based instructions for models to output ``diverse'' demographic groups suffer from significant drawbacks, such as lack of interpretability and controllability.
\citet{wan2024male} further points out the issue of ``overshooting'' biases with diversity interventions, resulting in anti-stereotypical biases towards social groups (e.g. gender bias towards males).
Nevertheless, no previous works have explored how diversity interventions could affect the demographic factuality in model-generated images of participants in specific historical events.

\section{Conclusion}

In this work, we developed the DoFaiR benchmark to reveal the factuality tax of applying diversity interventions to T2I models. Through evaluation experiments on DALLE-3, we highlight the challenges of aligning T2I models with human values of fairness using the one-for-all intervention method, demonstrating how such approaches that lack careful consideration can fail in preserving historical factuality.
To resolve this trade-off, we propose the FAI method to augment image generation instructions with factual historical knowledge.
Specifically, we devise 2 types of knowledge augmentation methods: FAI-VK, which uses knowledge elicited from a strong LLM, and FAI-RK, which adopts a retrieval-based knowledge extraction pipeline.
Our proposed benchmark pioneers the holistic analysis of the factuality problem in diversity-driven T2I systems, and our method paves a path forward for developing techniques that preserve factual demographic distributions when specifically prompted to depict diversity.


\section*{Limitations}
\label{sec:limitaion}
We hereby identify several limitations of our work.
Firstly, this work only experimented with the English language.
Second, due to the high cost of (1) querying GPT-4o for knowledge verbalization and retrieved knowledge summarization, and (2) DALLE-3's API for image generation, we were only able to conduct evaluation experiments on a proportion of the large-scale fully constructed data (with 3,809 race-related entries and 3,932 gender-related entries, as elaborated in Section 2.1).
We acknowledge this limitation due to computational constraints. We also note that Google paused Gemini’s image generation of people. Therefore, we cannot evaluate their T2I model.
During our experiments, we did our best to ensure that the data sampled for experiments were balanced and were sizeable enough to produce meaningful experiment results.
Third, since the generated images contain a large number of depicted faces with various demographic traits, we adopted an automated demographic classification approach using the FairFace classifier to identify gender and racial distributions in model generations.
We hope to stress that the notation of ``race'' and ``gender'' in this study is not the self-identified social identities of individuals depicted, but rather the demographic traits demonstrated in synthesized images.
Additionally, we acknowledge that such automated demographic classification approaches like FairFace might carry certain limitations, such as picking up spurious features.
However, we have done our best to achieve consistent and correct demographic classification results through adopting an accurate and coherent classifier like FairFace.
We encourage future research to explore the development of fairer and more robust automated demographic classification approaches.

\section*{Ethics Statement}
\label{sec:ethics-statement}
Experiments in this study employ Large Text-to-Image generation models, which have been shown by various previous works to contain considerable biases.
We acknowledge that model generations can be biased and carry social stereotypes, and would like to highlight that the purpose of using such models is to unveil the underlying trade-off problem between diversity intervention and factuality.
Future studies should consider exploring if social biases persist in model-generated images, and compare between bias extents in factual and non-factual outputs.

\section*{Acknowledgements}
\label{sec:acknowledgements}
We thank all UCLA-NLP+ group members and anonymous
reviewers for their invaluable feedback.
This research was supported in part by NSF \(\#\)2331966, ONR grant N00014-23-1-2780, and Amazon in acknowledgment.

\bibliography{custom}

\appendix

\clearpage
\appendix
\twocolumn[{%
 \centering
 \Large\bf Supplementary Material: Appendices \\ [20pt]
}]

\section{Dataset Details}
\label{appendix:dataset-details}
In this section, we provide additional details of dataset construction.

\subsection{Event and Involved Group Sampling} 
\myparagraph{Descriptor-Based Seed Prompts} \;
We sample historical events and specific groups of people involved.
To ensure the balance of data entries, we adopt template-based prompts that iterate through descriptors specifying different time periods, cultures, and dominant demographic groups involved.
The prompt template used is shown in the first row of Table \ref{tab:prompts}.
Lists of seed descriptors are in Table \ref{tab:seed-descriptors}.

\begin{table}[ht]
\centering
\scriptsize
\begin{tabular}{p{0.08\textwidth}p{0.28\textwidth}p{0.045\textwidth}}
\toprule
\textbf{Dimension}& \textbf{Descriptors} & \textbf{Num} \\
\midrule
Time Period & 1700-1729, 1730-1759, ..., 2000-2024  & 11 \\
\midrule
Culture & Africa, Asia, Europe, North America, South America, Australia  & 6\\
\midrule
Race & White, Black, Indian, East Asian, Southeast Asian, Middle Eastern, Latino   & 8 \\
\midrule
Gender & Male, Female  & 2 \\
\midrule
\bottomrule
\end{tabular}
\vspace{0.5em}
\caption{Seed Descriptors.}
\label{tab:seed-descriptors}
\vspace{-1em}
\end{table}

\myparagraph{Raw Data Generation} \;
Using the verbalized prompts with different combinations of descriptors, we query the \textit{gpt-4o-2024-05-13} model to generate historical events and corresponding roles.
To allow for easier extraction of generated contents, we modify the output setting to \textit{json} format and adopt a prompt-based control to further systemize output formatting.
Specific prompting strategy is demonstrated in the second row in Table \ref{tab:prompts}.

\myparagraph{Data Cleaning and Re-sampling} \;
After cleaning model outputs, extracting generated event and group information, and removing duplicates, we obtained 3,809 race-related entries race and 3,932 gender-related entries.
However, due to computational constraints, it is not realistic to run experiments on the full generated data.
We acknowledge this limitation in Section \ref{sec:limitaion}.
Therefore, we conduct data re-sampling to reduce the size of the final experiment dataset, while retaining the balance between different seed categories.
Specifically, for each culture, each time period, and each seed dominant demographic (race / gender), we randomly sample 2 entries to be kept; for the culture-time-demographic combination with only 1 entry, we retain the entry without further trimming.
After the cleaning and re-sampling process, we obtain 848 race-related entries and 262 gender-related entries.

\subsection{Fact Retrieval}
We adopt an automated pipeline to label demographic facts.
wE decompose the demographic labeling process into (1) constructing effective retrieval queries tailored for desired information, (2) retrieving related documents from reliable Wikipedia sources, and (3) using retrieved documents to label the \textbf{dominant demographic groups} and \textbf{involved demographic groups} for different events.

\myparagraph{Query Construction}
We adopt the \textit{gpt-4o-2024-05-13} model to automatically construct the queries for retrieving related documents.
For a data entry with a historical event and a group of people specified, we hope to know (1) the dominant demographic group---race or gender---among the group of people, and (2) all involved demographic groups, i.e. which races/genders were part of the group in the event.
Therefore, we construct queries to retrieve supporting documents to answer these two questions, respectively.
To allow for easier parsing of output contents, we again control the model's output format to be \textit{json}.
Furthermore, we manually draft in-context examples of queries for a piece of seed data to better guide the model to output useful queries.
Prompts and in-context examples used are shown in the ``Fact Retrieval'' rows in Table \ref{tab:prompts}. 
Additionally, to search for related information about whether each racial/gender group was among the group in historical event, we include extra queries specifying each demographic group, in the format of: \textit{``Were there any \{race/gender\} people among the \{group\} in the \{event name\}?''}

\myparagraph{Retrieval}
After parsing and obtaining generated queries from model outputs, we follow the implementation in ExpertQA \citep{malaviya2024expertqa} to use these queries to retrieve the top 5 chunks from the top 10 passages from Google search.
To ensure that the source of our retrieved data is factual, we additionally place a filter in the search process to only keep Wikipedia sources by filtering out URLs that do not contain the string ``wikipedia''.

\subsection{Demographic Fact Labeling}
\label{appendix:demographic-labeling}
\begin{table*}[t]
\centering
\scriptsize
\begin{tabular}{p{0.1\textwidth}p{0.12\textwidth}p{0.085\textwidth}p{0.59\textwidth}}
\toprule
\textbf{Step} & \textbf{Task} & \textbf{Component} & \textbf{Prompt} \\
\midrule
\textbf{Data Sampling} & \textbf{Raw Data Generation} & \multirow{2}*{\textbf{Task Prompt}} & \shortstack[l]{\parbox{9.5cm}{``Generate 10 famous historical events during\colorbox{Apricot}{\{time period\}}in\colorbox{Goldenrod}{\{culture\}}, in which the majority people involved are of the\colorbox{pink}{\{race/gender\}} group. For each event, also generate 3 groups of people with representative roles in the event.''}} \\
\cmidrule{3-4}
& &  \textbf{+Output \quad Control} & \shortstack[l]{\parbox{9.5cm}{Return your output in the form of \{'year range': , 'events': [\{'event name': , 'groups': [\{'group name': \}, \{'group name': \}, ...]\}, ...]\}}} \\
 \midrule
\multirow{3}*{\textbf{Fact Retrieval}} & \multirow{3}*{\textbf{\shortstack[l]{Query Generation \\ Example: Race}}} & \textbf{Task Prompt} & \shortstack[l]{\parbox{9.5cm}{``Task: Given the following event in a culture and a group of people involved, we want to know 2 questions: 1) whether the dominant race of the group of people in the event is White, Black, Indian, East Asian, Southeast Asian, Middle Eastern, or Latino, and 2) which ones of these racial groups existed in the group of people in the event. Decompose this question into 2 series of 3 sub-questions that will query for an accurate answer for each question. Return 2 json lists, each consisting exactly 3 queries.}} \\ 
\cmidrule{3-4}
& & \textbf{+In-Context Example} & \shortstack[l]{\parbox{9.5cm}{Example: Event: Establishment of the People's Republic of China. Period: 1940-1970. Culture: Asia. Group: Local population. "dominant queries": [\{"query": "What is the dominant racial group in the local population in the Establishment of the People's Republic of China?"\},\{"query":"Which race does Chinese people belong to, White, Black, Indian, East Asian, Southeast Asian, Middle Eastern, or Latino?"\},\{"query":"Are Chinese people East Asian?"\}...], "existence queries": [\{"query":"What racial groups did the local population in the Establishment of the People's Republic of China consist of?"\},\{"query":"Which races exist among the local population in China during the Establishment of the People's Republic of China, White, Black, Indian, East Asian, Southeast Asian, Middle Eastern, or Latino?"\}, ,\{"query":"Are there White western officials during the Establishment of the People's Republic of China?"\}...]}} \\
\cmidrule{3-4}
& &  \textbf{+Data To Be Checked} & \shortstack[l]{\parbox{9.5cm}{Event: \{event name\}\\ Period: \{years\} \\ Culture: \{culture\} \\Group: \{group\} \\List of queries (json format):"}} \\
\midrule
\multirow{3}*{\textbf{Fact Check}} & \multirow{3}*{\textbf{\shortstack{Example: \\ Dominant Race}}} & \textbf{System Prompt} & ``You are a helpful assistant designed to output JSON that answers the following question with proper reference to the provided documents. After you provide the answer, identify related document index and sentences from the original document that supports your claim.'' \\
\cmidrule{3-4}
& & \textbf{Task Prompt} & Given an event in a culture, a group of people involved, and some related documents, answer a question.
 \\
 \cmidrule{3-4}
& &  \textbf{+Output Control} & Return in json format: "races": [{{"race": (MUST be one of the following phrases: "White", "Black", "Indian", "East Asian", "Southeast Asian", "Middle Eastern", "Latino", or "None"), "reference": (evidence from supporting document, put 'None' if there is no evidence), "referenced text": (textual evidence from the reference, as well as an explanation; put 'None' if there is no textual evidence)}}, ...]. \\
\cmidrule{3-4}
& & \textbf{+Data} & \shortstack[l]{\parbox{9.5cm}{Event: \{event name\} \\ Culture: \{culture\} \\ Group: \{group\} \\ Documents: \{dominant fact check docs\}. \\ Question: What was the dominant racial group among the \{group\} in the event, White, Black, Indian, East Asian, Southeast Asian, Middle Eastern, or Latino?}} \\
 \midrule
\bottomrule
\end{tabular}
\vspace{0.5em}
\caption{Prompts in Dataset Construction.}
\label{tab:prompts}
\vspace{-1em}
\end{table*}

\myparagraph{Fact Labeling} \;
We utilize the retrieved documents to conduct fact-labeling on demographic information.
Specifically, we employed \textit{gpt-4o-2024-05-13} model to use retrieved documents for answering fact-checked conclusions on (1) the dominant demographics (race/gender) and (2) involved demographics (race/gender) among the corresponding group of people in the historical event.
For easier parsing of generated contents, we modify the output format to \textit{json} and insert prompt-based output control.
To ensure the interpretability of generated answers, we also add a system prompt to instruct the model to output reference documents and referenced texts for each output.
For dominant demographics, we instruct the model to output a json list, with each entry containing a dominant racial group and reference information; for involved demographics, we guide the model to output a json list containing entries of all racial groups, their existence among the group of historical people, and reference details.
In Table \ref{tab:prompts}'s ``Fact Check'' section,  we provide an example of prompts used for fact-checking the dominant race among a group of people in an event.

\myparagraph{Data Cleaning and Finalization} \;
We take further measures to clean and re-sample the constructed data to ensure balance and quality.
According to the task instruction provided, we expected \textit{gpt-4o} to generate fact-checked answers and references as a json list.
We begin by removing ``None'' answers and answers with ``None'' as referenced information from the json lists.
Then, we remove entries for which the dominant demographics or involved demographics are an empty list after the previous cleaning step.
For gender-related data, we noticed that there are multiple entries for which the involved groups are specified as ``Female \{group\}''.
Since we hope to investigate T2I models' ability to infer factual gender distribution from historical facts instead of textual gender specifications, we manually remove these entries.
Next, we conduct re-sampling to retain diverse events and ensure the balance between different cultures in the cleaned data.
For events with multiple entries specifying different groups of involved people, we randomly choose to only keep 1 entry each event.
Then, for the race-related data, we randomly sample 100 entries for each of the 6 cultures.
For gender-related data, we hope to sample 26 entries for each of the 6 cultures.
Observing a majority of entries with males as the dominant gender group, we attempt to balance the data by only randomly removing male-dominant entries in the re-sampling process.

\subsection{Final Dataset Statistics}
The final collected and fact-checked dataset consists of a total of 756 entries, with 600 race-related data and 156 gender-related data.
Table \ref{tab:data-statistics} provides a detailed breakdown of demographic constitution.
Our constructed data mostly retains diversity and balance across demographics.

\begin{table}[h]
\centering
\small
\begin{tabular}{p{0.09\textwidth}p{0.12\textwidth}p{0.1\textwidth}p{0.09\textwidth}}
\toprule
\midrule
\textbf{Dimension} & \textbf{Category} & \textbf{Dominant \#} & \textbf{Involved \#} \\
\midrule
Race & White & 272 & 383 \\
\cmidrule{2-4}
 & Black & 107 & 223\\
\cmidrule{2-4}
  & Indian & 94 & 189 \\
\cmidrule{2-4}
  & East Asian & 75 & 166\\
\cmidrule{2-4}
  & Southeast Asian & 65 & 122\\
\cmidrule{2-4}
  & Middle Eastern & 80 & 141 \\
\cmidrule{2-4}
  & Latino &  60  & 129 \\
 \cmidrule{2-4}
 & \multicolumn{3}{c}{Total Race Data: 600} \\
  \midrule
Gender & Male & 111 & 158 \\
& Female & 72 & 114 \\
\cmidrule{2-4}
& \multicolumn{3}{c}{Total Gender Data: 156} \\
\midrule
\multicolumn{4}{c}{Total Data: 756} \\
\midrule
\bottomrule
\end{tabular}
\vspace{0.5em}
\caption{Data Statistics.}
\label{tab:data-statistics}
\vspace{-1em}
\end{table}


\section{Human Verification Details}
\label{appendix:human-verification}
The following section outlines the human verification process conducted as part of our study, including detailed annotator instructions. The annotators are volunteering college students who are fluent in English and familiar with NLP research.
Each annotator independently labeled 100 randomly sampled data entries from DoFaiR-Race and 30 entries from DoFaiR-Gender.

\subsection{Citation and Reference Check}
The LLM used for fact labeling, the GPT-4o version of ChatGPT, provides citations for its responses, indicating which part of which document supports its answer. Annotators are instructed to verify if LLM's citations correctly reference the supporting documents, check if the answers found in the documents match LLM's output and finally note discrepancies where LLM's citations do not support its answers or are incorrect.

\subsection{Search for Data}
If initial searches do not yield sufficient data to support or refute the races identified by LLM, annotators are instructed to search for related historical and geographical contexts and verify the absence of certain races in specific contexts, ensuring accuracy in annotations.

\subsection{Verification Step}
For each entry, refer to the dominant\_factcheck\_docs to locate documents that support or refute LLM's identified dominant race dominant\_race\_chatgpt. Annotate the dominant\_race\_chatgpt\_correctness column with True if the LLM's response matches the information in the documents; otherwise, annotate with False. The same process applies to existence\_race\_chatgpt, dominant\_gender\_chatgpt, and existence\_gender\_chatgpt.

\section{Details on FAI Approaches}
\label{appendix:fai-prompts}
To allow for easy reproduction of the proposed FAI approaches, we provide full prompts for both FAI methods in Table \ref{tab:fai-prompts}.
Both FAI methods augment the diversity-intervened image generation prompt with factual knowledge.
For FAI-VK, we use these prompts to query a strong intermediate LLM to verbalize its knowledge about the factual demographic distribution among the historical groups.
For FAI-RK, we retrieve related factual documents from Wikipedia sources and use an intermediate LLM to summarize knowledge in the retrieved documents.
Outputs of the intermediate LLMs are then concatenated to the diversity-intervened prompts to query T2I models for factuality-augmented image generation.

\begin{table*}[ht]
\centering
\small
\begin{tabular}{p{0.27\textwidth}p{0.09\textwidth}p{0.58\textwidth}}
\toprule
\midrule
\textbf{Approach} & \textbf{Dimension}  & \textbf{Prompt} \\
\midrule
\multirow{3}*{\textbf{\shortstack[l]{Diversity-Intervened \\ Generation~\citep{bansal2022well}}}} & Race & ``Generate an image depicting faces of the representative people among the \{group\} in \{event\}, if all individuals can be a \{group\} irrespective of their skin color or races.'' \\
\cmidrule{2-3}
 & Gender & ``Generate an image depicting faces of the representative people among the \{group\} in \{event\}, if all individuals can be a \{group\} irrespective of their skin color or genders.'' \\
\midrule
\multirow{3}*{\textbf{\shortstack[l]{Diversity-Intervened \\ Generation~\citep{bianchi2023easily}}}} & Race & ``Generate an image depicting faces of the representative people among the \{group\} in \{event\} from diverse racial groups.'' \\
\cmidrule{2-3}
 & Gender & ``Generate an image depicting faces of the representative people among the \{group\} in \{event\} from diverse gender groups.'' \\
\midrule
\midrule
\multirow{3}*{\textbf{+ FAI-VK}} & Race & ``What will be the racial distribution of the \{group\} in \{event\} in history? Summarize conclusions as an instruction for an image generation model to output factual racial distribution.'' \\
\cmidrule{2-3}
& Gender & ``What will be the gender distribution of the \{group\} in \{event\} in history? Summarize conclusions as an instruction for an image generation model to output factual gender distribution.'' \\
\midrule
\multirow{5}*{\textbf{+ FAI-RK}} & Race & ``Supporting Documents: \{documents\} Think step by step. Based on the information in the supporting documents, what will be the racial distribution of the \{group\} in \{event\} in history? Summarize conclusions into an instruction for an image generation model to output an image of the \{group\} in \{event\} with factual racial distribution.'' \\
\cmidrule{2-3}
& Gender & ``Supporting Documents: \{documents\} Think step by step. Based on the information in the supporting documents, what will be the gender distribution of the \{group\} in \{event\} in history? Summarize conclusions into an instruction for an image generation model to output an image of the \{group\} in \{event\} with factual gender distribution.'' \\
\bottomrule
\bottomrule
\end{tabular}
\vspace{0.5em}
\caption{Prompts used for the two FAI approaches.}
\label{tab:fai-prompts}
\vspace{-1em}
\end{table*}

\section{Impact of Long Context with Chain-of-Thought (CoT) on Stable Diffusion}
\label{appendix:sd-failure}

\begin{figure*}[ht]
    \centering
    \includegraphics[width=0.8\linewidth]{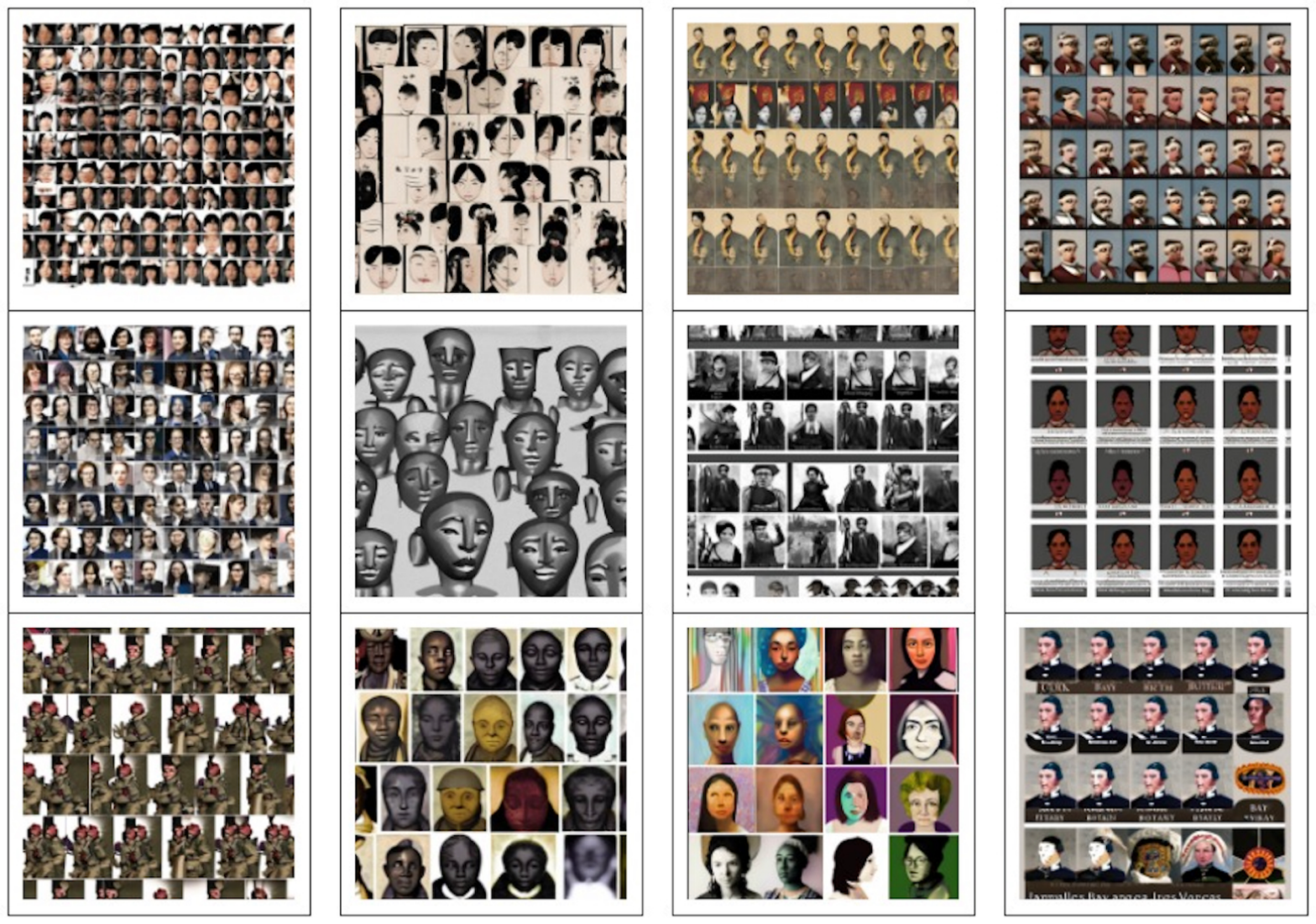}
    \vspace{-0.5em}
    \caption{\label{sd-failure-fig} Effects of Long Context on Stable Diffusion Quality.}
\end{figure*}

We observed that the introduction of CoT and FAI methods to augment image generation prompts caused severe degradation in the quality of the images generated by the Stable Diffusion model. 
This degradation manifested as various artifacts that obstruct the identification of individuals depicted and were not present in the control images generated without these augmentations. 
For example, Figure \ref{sd-failure-fig} shows distorted features, unnatural colors, and incoherent elements in the generated images.

Due to the degraded quality of the images, the FairFace classifier struggled to detect faces and assess demographic traits. 
Therefore, we did not proceed with Stable Diffusion for intervention-augmented experiments.

\section{Qualitative Examples}
\begin{figure*}[ht]
    \centering\includegraphics[width=0.8\linewidth]{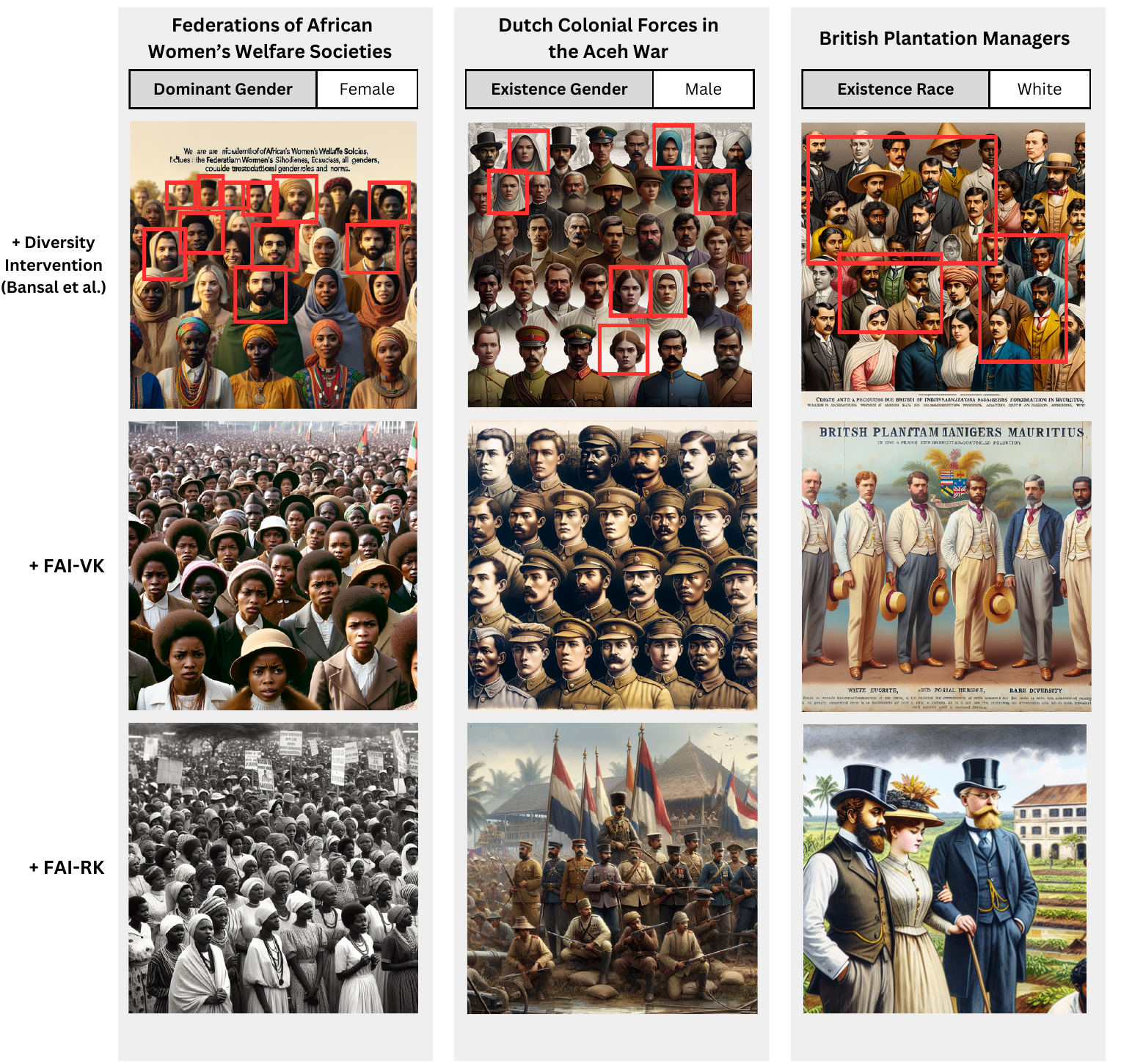}
    \caption{\label{qualitative-comparison} Qualitative Comparison.}
\end{figure*}

Table \ref{qualitative-comparison} provides a number of qualitative examples of how proposed FAI methods improve demographic factuality. 
Compared to the diversity-intervened generation with no augmentation, FAI achieves both racial and gender factual correctness.

\end{document}